\documentclass[letterpaper]{article} 
\usepackage{arXiv_aaai24}  
\usepackage{times}  
\usepackage{helvet}  
\usepackage{courier}  
\usepackage[hyphens]{url}  
\usepackage{graphicx} 
\usepackage{amsmath}
\usepackage{natbib}  
\usepackage{caption} 
\usepackage{algorithm}
\usepackage{algorithmic}
\usepackage{booktabs}
\usepackage{multirow}
\usepackage{multicol}
\usepackage{diagbox}
\usepackage{amssymb}
\usepackage{amsfonts}

\usepackage{siunitx}
\usepackage{booktabs}
\usepackage{array}
\usepackage{subfigure}
\usepackage{graphicx}
\usepackage{marvosym}
\usepackage{colortbl}
\usepackage{pifont}

\usepackage{newfloat}
\usepackage{listings}
\DeclareCaptionStyle{ruled}{labelfont=normalfont,labelsep=colon,strut=off} 
\floatstyle{ruled}
\newfloat{listing}{tb}{lst}{}
\floatname{listing}{Listing}
\title{MM-Point: Multi-View Information-Enhanced Multi-Modal Self-Supervised 3D Point Cloud Understanding}
\author{
    Hai-Tao Yu\textsuperscript{\rm 1,3}, 
    Mofei Song\textsuperscript{\rm 2,3} \thanks{Corresponding Author}
}
\affiliations{
    \textsuperscript{\rm 1} School of Cyber Science and Engineering, Southeast University, Nanjing, China\\
    \textsuperscript{\rm 2} School of Computer Science and Engineering, Southeast University, Nanjing, China\\
    \textsuperscript{\rm 3} Key Laboratory of New Generation Artificial Intelligence Technology and Its Interdisciplinary Applications (Southeast University), Ministry of Education, China \\
    
     $\{$yuht, songmf$\}$@seu.edu.cn
}

\usepackage{bibentry}

\begin{document}

\maketitle

\begin{abstract}
In perception, multiple sensory information is integrated to map visual information from 2D views onto 3D objects, which is beneficial for understanding in 3D environments. But in terms of a single 2D view rendered from different angles, only limited partial information can be provided.
The richness and value of Multi-view 2D information can provide superior self-supervised signals for 3D objects. In this paper, we propose a novel self-supervised point cloud representation learning method, \textbf{MM-Point}, which is driven by \textit{intra-modal} and \textit{inter-modal} similarity objectives. 
The core of MM-Point lies in the Multi-modal interaction and transmission between 3D objects and multiple 2D views at the same time. In order to more effectively simultaneously perform the consistent cross-modal objective of 2D multi-view information based on contrastive learning, we further propose \textbf{\textit{Multi-MLP}} and \textbf{\textit{Multi-level Augmentation}} strategies. 
Through carefully designed transformation strategies, we further learn Multi-level invariance in 2D Multi-views. 
MM-Point demonstrates state-of-the-art (\textbf{SOTA}) performance in various downstream tasks. For instance, it achieves a peak accuracy of $92.4\%$ on the synthetic dataset ModelNet40, and a top accuracy of $87.8\%$ on the real-world dataset ScanObjectNN, comparable to fully supervised methods. Additionally, we demonstrate its effectiveness in tasks such as few-shot classification, 3D part segmentation and 3D semantic segmentation.
\end{abstract}

\begin{figure}[t]
  \centering
  \includegraphics[width=\columnwidth]{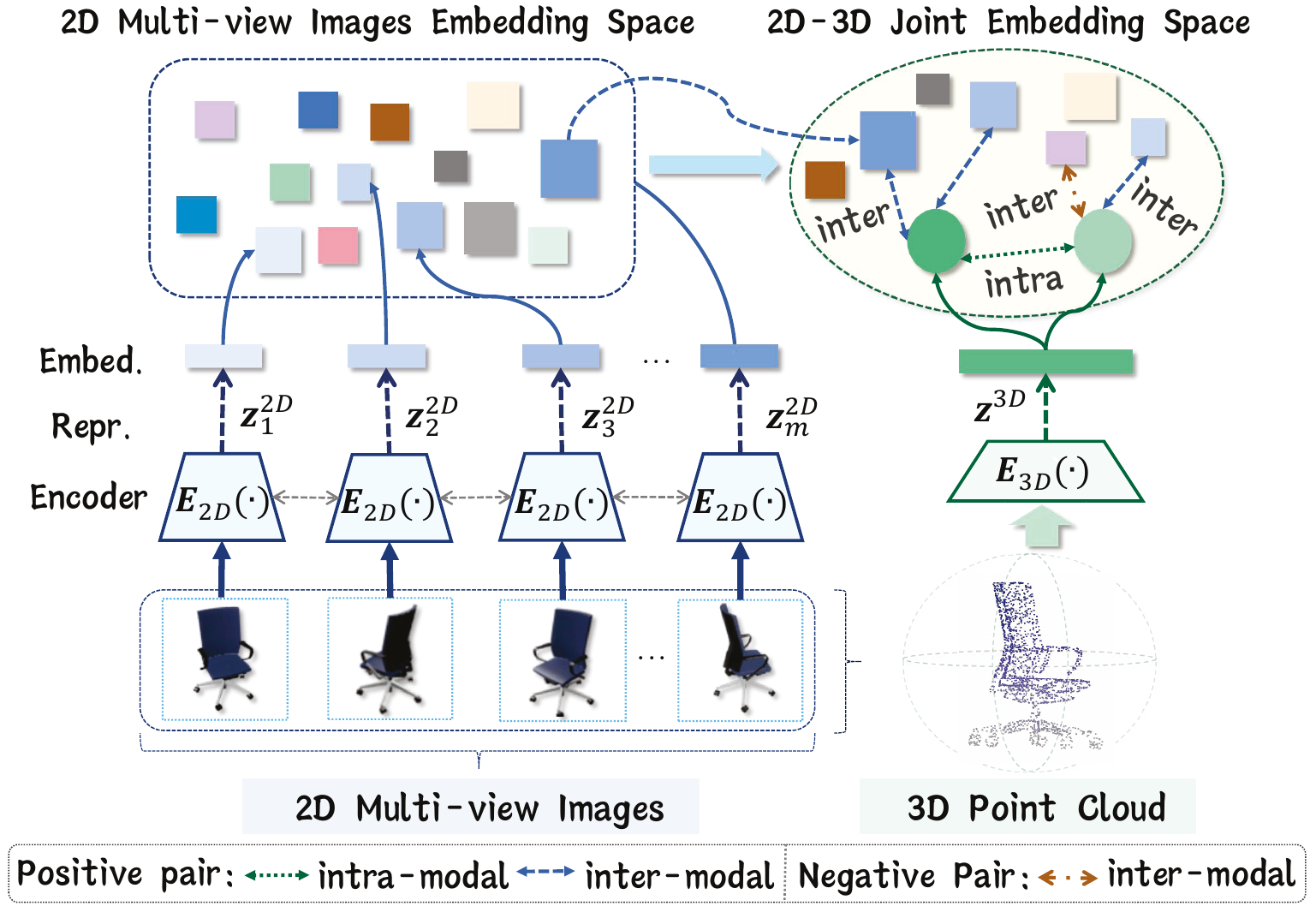}
  \caption{Schematic of MM-Point multi-modal contrastive learning. Given 2D multi-views rendered from a 3D object, the model distinguishes the views of the same object from those of different objects in the embedding space, enabling self-supervised learning of deep representation. }
  \label{fig:intro}
\end{figure}

\section{Introduction}
\label{sec:1}
In recent years, the demand for 3D perception technologies has surged in the real world. As a fundamental 3D representation, point cloud learning plays a crucial role in numerous tasks, including 3D object \textit{classification}, \textit{detection}, and \textit{segmentation}. However, the cost of point cloud annotation is high, and 3D scans with labels are usually scarce in reality. To address these challenges, many studies have turned their attention to self-supervised methods.

Interestingly, while 3D point clouds can be obtained by sampling, 2D multi-view images can be generated by rendering. Some recently proposed methods \cite{yang2020pointflownet, huang2020s3dis} have focused on generating high-quality point cloud representations using multi-modal data.

This naturally raises a question: Can we better facilitate the understanding of 3D point cloud representations by leveraging the abundant information hidden in 2D Multi-views? We reconsider the contrast paradigm in 2D-3D, hypothesizing that 3D objects and rendered 2D Multi-views share mutual information. 
Our primary motivation is to view each 2D view as a unique pattern that is useful for guiding 3D objects and improving their representations, as each 2D view observes different aspects of 3D objects and representations from different angles are distinctive. From another perspective, we encourage the 3D-2D relationship to be consistent across different-angle views, \textit{i.e.} the similarity correspondence between the 2D images and 3D point cloud in one view should also exist in another view.

Due to the distinct nature of 3D representations from 2D visual information, a simple alignment between these two types of representations may lead to limited gain or even negative transfer in multi-modal learning. Therefore, we propose a novel 3D point cloud representation learning framework and a multi-modal pre-training strategy between 3D and 2D, namely \textbf{MM-Point} (as shown in Fig.\ref{fig:intro}), which transfers the features extracted from 2D multi-views into 3D representations. Specifically, \textit{intra-modal} training is used to capture the intrinsic patterns of 3D augmented data samples. The \textit{inter-modal} training scheme aims to learn point cloud representations by accepting 2D-3D interactions. Meanwhile, considering the differences in multi-modal data properties, 
We further ponder: how to effectively simultaneously transfer multiple 2D view information from different angles to 3D point cloud objects in a self-supervised manner$?$

Extending this, we propose a \textbf{Multi-MLP} strategy to construct multi-level feature learning between each 2D view and 3D object, thus the consistency goal is set to contrast between 2D multi-views and 3D objects across different feature spaces. Such architectural design not only extracts shared information in 2D-3D multi-modal contrast, 
but also preserve specific information in multiple 2D views at the same time, 
thereby the overall consistency goal could better extract semantic information between 3D point clouds and as many different 2D views as possible, resulting in improved 3D representations.

Additionally, treating each angle of 2D view equally during training with the same type of augmentation transformation may lead to a suboptimal representation for downstream tasks in multimodal contrast. We suggest a \textbf{Multi-level Augmentation} strategy based on multi-view, integrating rendered 2D multi-view information with augmentation information. Moreover, we control the strength of all augmentation modules, ensuring the mutual information between 3D point cloud and 2D image augmentation pairs remains low and within a certain range, thus enabling the model to gradually accumulate more complex information during learning.

\begin{figure*}[t]
  \centering
  \includegraphics[width=0.91\textwidth]{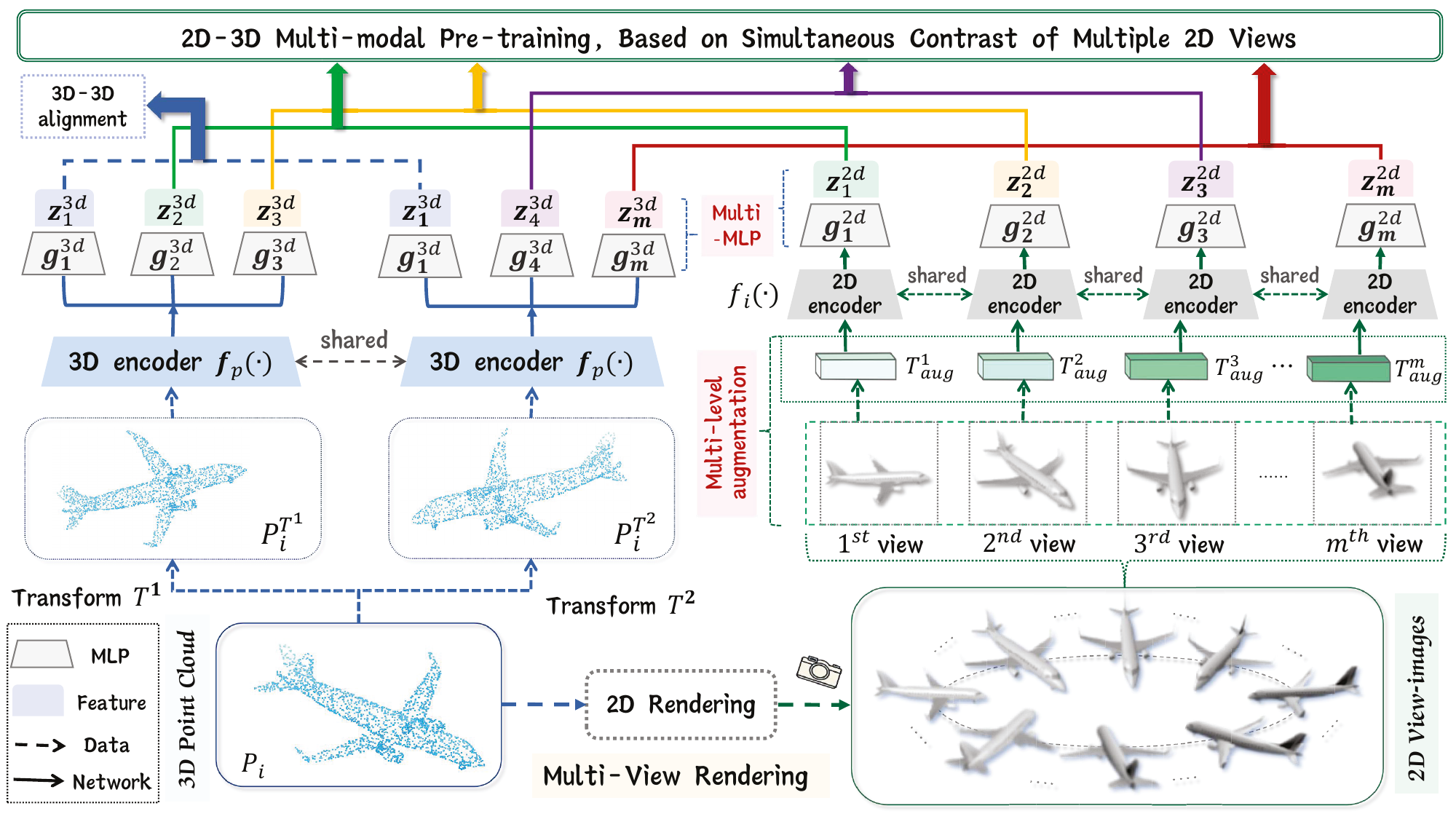}
  \caption{Schematic of the MM-Point architecture. 
  MM-Point carries out intra-modal self-supervised learning in the 3D point cloud (blue path) and cross-modal learning between 2D and 3D (other color paths), aligning 2D multi-view features with 3D point cloud features. To better utilize the information from the 2D multi-views, MM-Point introduces two strategies: 1) Multi-MLP: constructing a multi-level feature space; 2) Multi-level augmentation: establishing multi-level invariance.}
  \label{fig:architect}
\end{figure*}

To evaluate our proposed 2D-3D multi-modal self-supervised learning framework, we assess the performance of MM-Point across several downstream tasks. The learned 3D point cloud representation can be directly transferred to these tasks. For 3D object shape classification, we performed on both synthetic dataset ModelNet40 and real-world object dataset ScanObjectNN, achieving state-of-the-art performance, surpassing all existing self-supervised methods. 
Furthermore, part segmentation and semantic segmentation experiments validate MM-Point's capability to capture fine-grained features of 3D point clouds.

In summary, our research contributions are as follows:

\begin{itemize}
    \item We introduce \textbf{MM-Point}, a novel 3D representation learning scheme based on 2D-3D multi-modal training. 
    \item Our research applies multi-modal contrastive learning to multi-view settings, maximizing shared mutual information between different 2D view and the same 3D object. 
    \item We propose \textbf{Multi-MLP} and \textbf{Multi-level Augmentation} strategies, thereby ensuring more effective learning of 3D representation in multi-modal contrast under a self-supervised setting, achieving effective pre-training from 2D multi-views to 3D objects.
    \item \textbf{MM-Point} demonstrates remarkable transferability. The pre-trained 3D representations can be directly transferred to numerous downstream tasks, achieving state-of-the-art performance in extensive experiments.
\end{itemize}

\section{Related Work}
\label{sec:2}
\subsection{Self-supervised Point Cloud Learning}
\label{sec:2.1}
Unsupervised learning for point cloud understanding can be broadly divided into generative or discriminative tasks based on the proxy tasks. Generative models learn features by self-reconstruction, as exemplified by methods like JigSaw \cite{qi2018jigsaw} .
Furthermore, Point-BERT \cite{yu2022point} predicts discrete labels, while Point-MAE \cite{pang2022masked} randomly masks patches of the input point cloud and reconstructs the missing parts. However, these methods are computationally expensive. On the other hand, discriminative methods predict or discriminate the enhanced versions of the inputs. Our work learns point cloud representations based on this approach. 
Recent works \cite{you2021h3dnet, sun2021seedcon} have explored contrastive learning for point clouds following the success of image contrastive learning. PointContrast \cite{xie2020pointcontrast} is the first unified framework investigating 3D representation learning with a contrastive paradigm. 
Compared to these works, we investigate contrastive pre-training of point clouds from a new perspective, leveraging the semantic information hidden in 2D multi-views to design a multi-modal learning network, and thus enhancing the representational capacity of 3D point clouds.

\subsection{Multi-modal Representation Learning}
\label{sec:2.2}
Recent studies on self-supervised learning have leveraged the multi-modal attributes of data \cite{caron2020swav,wang2020vision}
, with a common strategy being the exploration of natural correspondences across differing modals, emphasizing the extraction of cross-modal shared information. For instance, in the field of visual-textual multi-modalities, large-scale image-text pairs \cite{li2020visualbert}
 have been pre-trained, enabling these models to be applicable for numerous downstream tasks. 
 A few works have integrated point cloud representations with other modalities, such as voxels \cite{shi2020pv,shi2020pv++}
 or multi-view images \cite{shi2019pointrcnn,qian2020pointgmm}
. Prior3D \cite{liu2020prior3d}
 proposed a geometric prior contrastive loss to enhance the representation learning of RGB-D data. Tran \textit{et al.} \cite{tran2022self} employ self-supervision through local correspondence losses and global losses based on knowledge distillation. Of note, CrossPoint \cite{afham2022crosspoint}, most related to our work, learns 2D-3D cross-modal representations via contrastive learning, concentrating on shared features across different modes. In comparison, our approach diverges significantly. 
 MM-Point is able to utilize information from multiple views in 2D space simultaneously for self-supervision of 3D point cloud features. 
 By contrasting with Multi-views simultaneously, it better facilitates the maximization of mutual information shared across multi-modalities, leading to more robust and enhanced performance in various downstream tasks.

\section{The Proposed Method}
\label{sec:3}
The proposed MM-Point multi-modal pre-training architecture is depicted in Fig. \ref{fig:architect}. It comprises two contrasting training strategies: \textit{intra-modal} point cloud contrastive learning and \textit{inter-modal} 2D-3D contrast, each accompanied by different types of loss functions. The \textit{inter-modal} training scheme enhances 3D point cloud features by interacting with rendered 2D multi-views, while the point cloud representation focuses on the shared mutual information among multiple 2D images at the same time, inherently boosting the diversity of multi-modal contrastive learning. Furthermore, we employ a Multi-MLP strategy to project multi-view and multi-modal features. Finally, we put forward a Multi-level augmentation invariance strategy based on 2D multi-views information rendered from the same 3D object.

\subsection{Intra-modal and Inter-modal Alignment} 
\label{subsection:3.1}

We propose to learn a encoder for aligning 3D point cloud features with visual features of 2D views. The pretraining process jointly handles the alignment of multiple modalities.

For each point cloud $P_i$, we obtain two variants, $P_i^1$ and $P_i^2$, through augmentation operations. We then encode the augmented point clouds separately into the feature space $F_i^1, F_i^2 \in \mathbb{R}^{n \times d}$. The projected features are subsequently mapped into the latent space, producing the representations $z_i^1$ and $z_i^2$. By performing contrastive loss, we enforce that the distance between feature representations of the same object is smaller than the distance between different objects.

MM-Point aims to learn two objectives through cross-modal alignment: $f_P($.$)$ and $f_I($.$)$. 
Cross-modal contrast seeks to minimize the distance between point clouds and the corresponding rendered 2D images while maximizing the distance from other images.

Given a sample pair $\left(P_i, I_i\right)$, where $P_i$ and $I_i$ represent the embeddings of the 3D point cloud and its rendered 2D image description, respectively. For each sample $P_i$ in the mini-batch $\mathcal{M}$, the negative sample set $\mathrm{N}_i$ is defined as $\mathrm{N}_i=\left\{I_j \mid \forall I_j \in \mathcal{M}, j \neq i\right\}$. The corresponding cross-modal contrastive loss on $\mathcal{M}$ is as follows:

\begin{equation}
\resizebox{0.9\columnwidth}{!}{$
\textit{ Loss }_{\textit{inter }}=\mathbb{E}_{i \in \mathcal{M}}\left[-\log \frac{\exp \left(f_P\left(P_i\right)^T f_{\mathrm{I}}\left(I_i\right)\right)}{\exp \left(f_P\left(P_i\right)^T f_{\mathrm{I}}\left(I_i\right)\right)+\sum_{I_j \in \mathrm{N}_i} \exp \left(f_P\left(P_i\right)^T f_I\left(I_j\right)\right)}\right]
$}
\end{equation}

In addition, we map cross-modal features to different spaces and decouple training within and across modalities. 
We design projection heads with larger dimensions for cross-modal contrastive learning. 

\begin{figure}[t]
  \centering
  \includegraphics[width=\columnwidth]{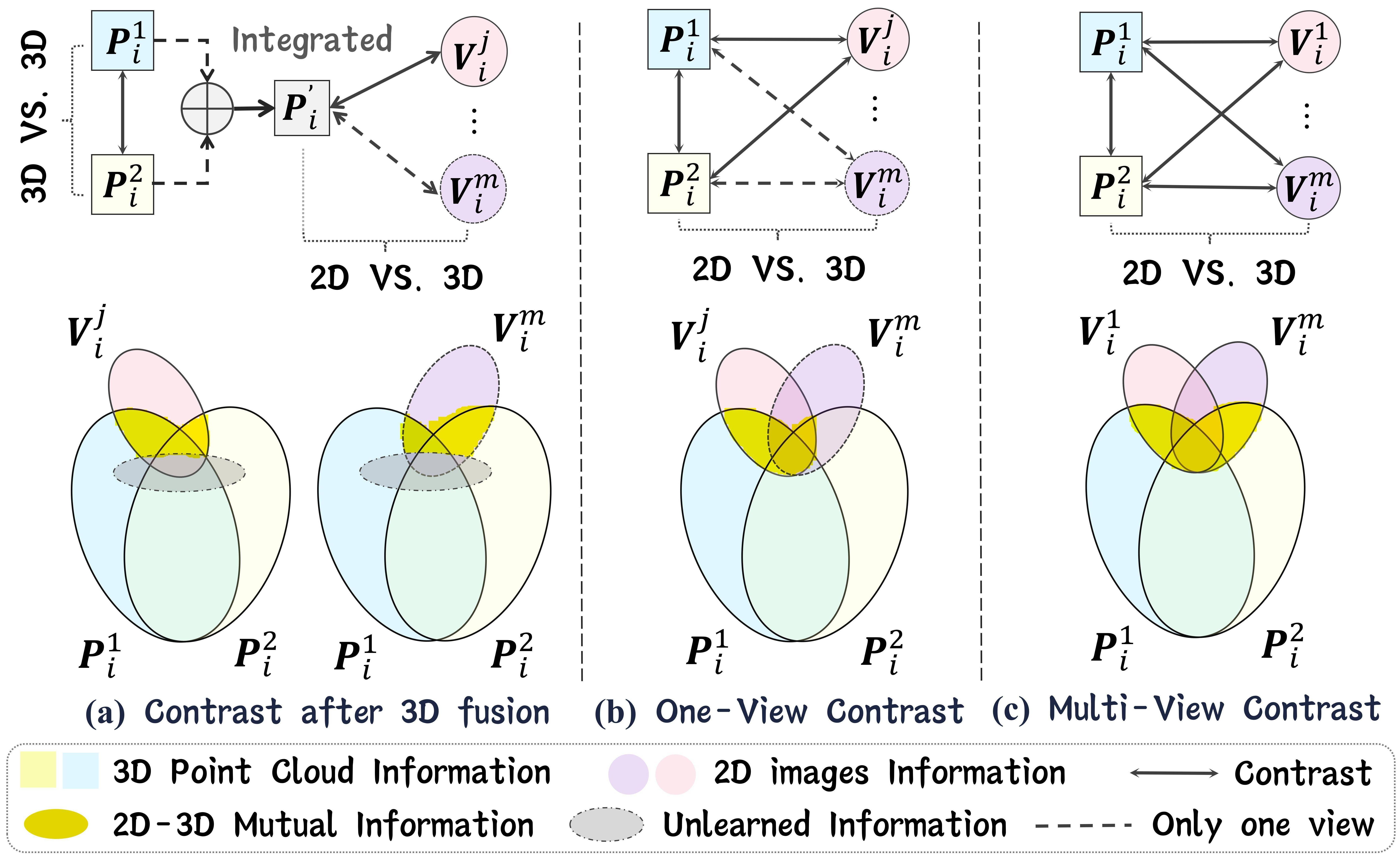}
  \caption{Mutual Information plot for the modal comparison between 3D and 2D. The yellow area signifies mutual information. The figure illustrates the impact of different strategies (a-c) on the contribution to mutual information.}
  \label{fig:CMI}
\end{figure}

\subsection{Multi-Modal Learning Based on Multi-View}\label{sec:3.2}

In this section, we describe the contrast between 3D point clouds and 2D multi-view images. By pairing multiple 2D images from different angles with a 3D object, this collaboration scheme benefits the model. 

\subsubsection{Rethinking the Contrast between 3D Object and 2D Multi-view}
For 3D point clouds and their corresponding rendered 2D multi-view images, we simply fix the 3D point clouds and enumerate positive and negative samples from the rendered 2D view images. Suppose the loss $\mathcal{L}_{\text {contrast }}^{P_i V_i}$ treats the 3D point cloud $P_i$ as an anchor and enumerates the 2D rendered views $V_i$. Symmetrically, we can obtain the loss by anchoring on the 2D $V_i$ and enumerating the 3D $P_i$. Then, we sum them up as the overall contrastive loss:$\quad \mathcal{L}\left(P_i, V_i\right)=\mathcal{L}_{\text {contrast }}^{P_i, V_i}+\mathcal{L}_{\text {contrast }}^{V_i, P_1}$.

Combining the theoretical proof in \textit{CMC} \cite{tian2019contrastive}, minimizing the loss should be equivalent to maximizing the lower bound of  $I\left(z_i ; z_j\right)$ , that is: $I\left(z_i ; z_j\right) \geq$ $\log (k)-\mathcal{L}_{\text {contrast }}$. In this case, $z_i$ and $z_j$ represent the latent representations of the point cloud and image, respectively, while $\mathrm{\textit{k}}$ denotes the number of negative sample pairs. 
Besides, research \cite{hjelm2019mutual,chen2021normalized} has shown that the boundaries of $I\left(z_i ; z_j\right)$ 
may not be clear, and finding a better mutual information estimator is more important. We consider the 3D point cloud and rendered multi-angle 2D views to construct all possible relationships between different 2D views and 3D point clouds. By involving all pairs, our optimized objective function is:
\begin{equation}
\mathcal{L}_F=\sum_{1 \leq j \leq M} \mathcal{L}\left(P_1, V_j\right)+\mathcal{L}\left(P_2, V_j\right)
\end{equation}

When learning 2D-3D contrast, the mutual information will change proportionally with the number of 2D views. When the view count reaches a certain level, multi-modal mutual information will reach a stable level. In Fig.\ref{fig:CMI}, the visualization demonstrates that contrasting 2D Multi-views with 3D point clouds enables the point cloud features to capture more mutual information between different 2D views.

\subsubsection{Multi-modal Contrastive based on 2D Multi-view}
MM-Point aims to maximize the mutual information between the differently augmented 3D point clouds and 2D views from various angles in the same scene. Naturally, too few or too many rendered 2D views exhibit poor mutual information performance, while an optimal position exists in between. We randomly sample $m$ 2D rendered views from different angles. The value of $m$ will be introduced in experiment.

Let $\mathcal{D}$ represent the pretraining dataset $\mathcal{D}=\left\{P_i,\left\{\mathbf{I}_{i j}, M_{i j}\right\}_{j=1}^m\right\}_{i=1}^n$, where $n$ denotes the number of 3D objects and $m$ represents the number of 2D views. Let $P_i$ be the 3D point cloud of the $i$-th object, and $\mathrm{\textit{I}}_{i j}$ denotes the 2D image of the $j$-th view of the $i$-th 3D object. For the multi-view 2D images $\mathrm{\textit{I}}_{i j}$, we extract features $\mathbf{H}_{i j}^I \in \mathbb{R}^{1 \times C}$, where $j \in\{1, m\}$. These 2D image features correspond to the feature $\mathbf{Z}_i^P$ of the $i$-th 3D object. Building on the 2D-3D cross-modal contrastive scheme, we further extend it and design a cumulative loss $\textit{Loss}_{\textit{inter-plus }}$: 



\begin{equation}
\resizebox{0.9\columnwidth}{!}{$
\begin{gathered}
J_{\textit{inter}}^i(P, I) = \sum_{k=1}^n \exp \left(\operatorname{sim}\left(\mathbf{Z}_i^P, \mathbf{Z}_k^P\right) / \tau\right) + \exp \left(\operatorname{sim}\left(\mathbf{Z}_i^P, \mathbf{H}_k^I\right) / \tau\right), \\
\mathcal{L}_{\textit{inter}}^i(P, I) = -\log \frac{\exp \left(\operatorname{sim}\left(\mathbf{Z}_i^P, \mathbf{H}_i^I\right) / \tau\right)}{J_{\textit{inter}}^i(P, I) - \exp \left(\operatorname{sim}\left(\mathbf{Z}_i^P, \mathbf{Z}_i^P\right) / \tau\right)}, \\
\textit{Loss}_{\textit{inter-plus}} = \frac{1}{2 n} \sum_{i=1}^n \sum_{j=1}^m \left(\mathcal{L}_{\textit{inter}}^i\left(P, I_j\right) + \mathcal{L}_{\textit{inter}}^i\left(I_j, P\right)\right)
\end{gathered}
$}
\end{equation}

\begin{figure}[t]
  \centering
  \includegraphics[width=0.8\columnwidth]{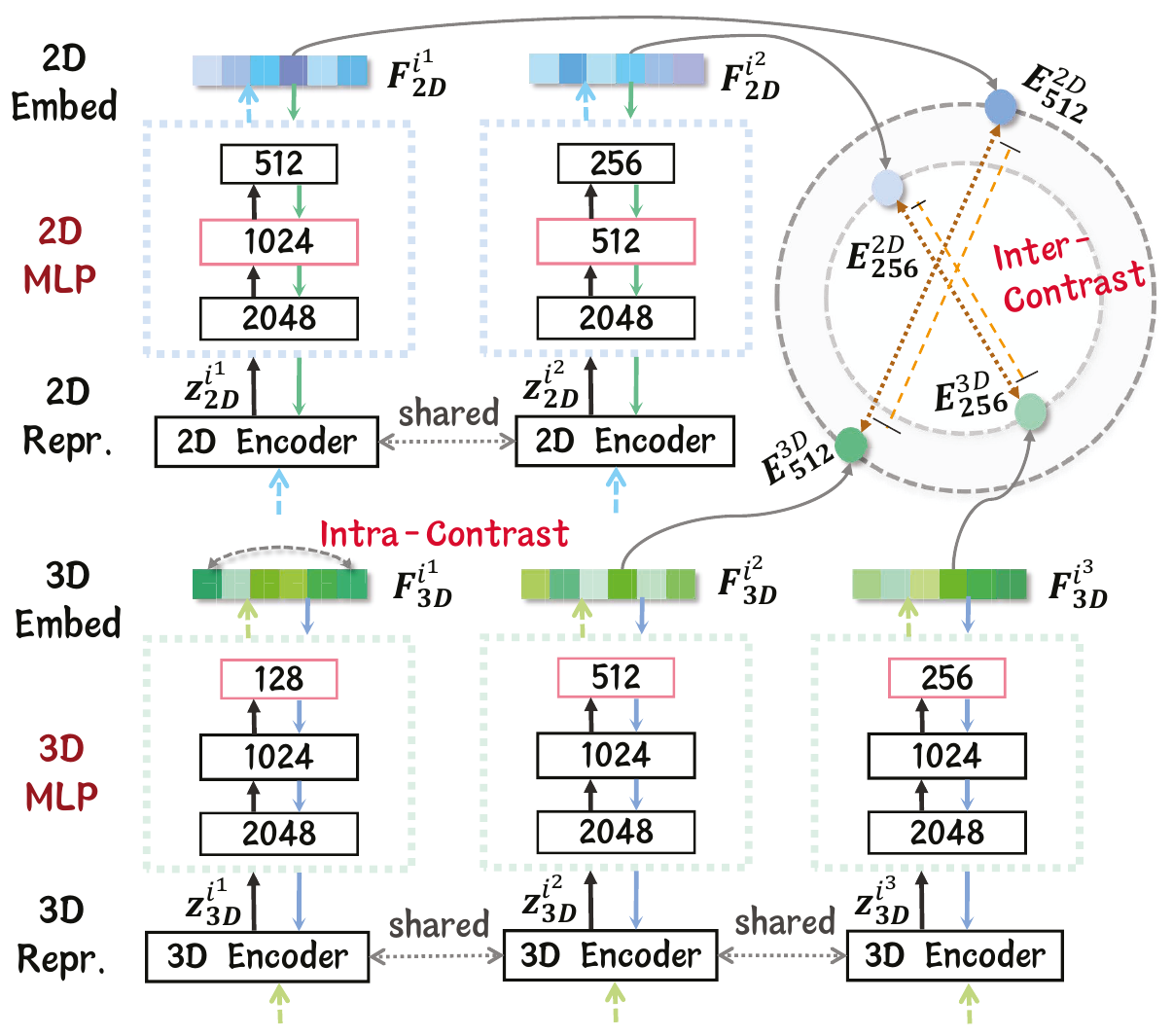}

  \caption{Schematic of the Multi-MLP strategy. Multi-modal learning enhances point cloud representation through 2D-3D interaction. Multiple 2D features and 3D features are contrasted through a multi-level feature space and adjusted via multi-path backpropagation.}
  \label{fig:MLP}
\end{figure}

\subsubsection{Multi-modal Contrastive based on Multi-MLP}
To alleviate the consistency conflicts between different 2D views and 3D object, we construct a multi-level feature learning strategy based on Multi-MLP. The multi-modal objective is then achieved by contrasting 2D multi-view and 3D objects in different dimensional feature spaces, as depicted in Fig.\ref{fig:MLP}.

Building upon the modality-specific and cross-modal training paths designed, for multiple 2D views, we further stack different MLPs on the encoder $\left\{\boldsymbol{f}^P, \boldsymbol{f}^I\right\}$, mapping them to distinct feature spaces. The mappings for the 3D encoder and 2D encoder are as follows:
\begin{equation}
\begin{gathered}
F\left(\left\{\mathbf{G}_H\right\}_{H=1}^m ; \boldsymbol{f}^P\right) ; \\
F\left(\left\{\mathbf{G}_H\right\}_{H=1}^m ; \boldsymbol{f}^I\right)
\end{gathered}
\end{equation}

Here, $\left\{\mathbf{G}_H\right\}_{H=1}^m$ denotes the additional projection heads of the MLP layers of the $m$ 2D multi-views, each with different output dimensions. 
The output feature dimension of \textit{cross-modal} projection heads exceeds that of \textit{intra-modal} outputs. 
The feature extraction process for the $j$-th 2D view corresponding to the $i$-th 3D object is as follows:

\begin{equation}
\resizebox{0.85\columnwidth}{!}{$
\left\{\mathbf{F}^H\right\}_{H=j}=\left\{\mathbf{G}_H\right\}_{H=j}\left(\mathbf{H}_{i j}^I\right)=\left\{\mathbf{G}_H\right\}_{H=j}\left(f_I\left({I}_{{ij}}\right)\right)
$}
\end{equation}

\subsection{Multi-level Augmentation Invariance}\label{sec:3.3}
\begin{figure}[h]
  \centering
  \includegraphics[width=\columnwidth]{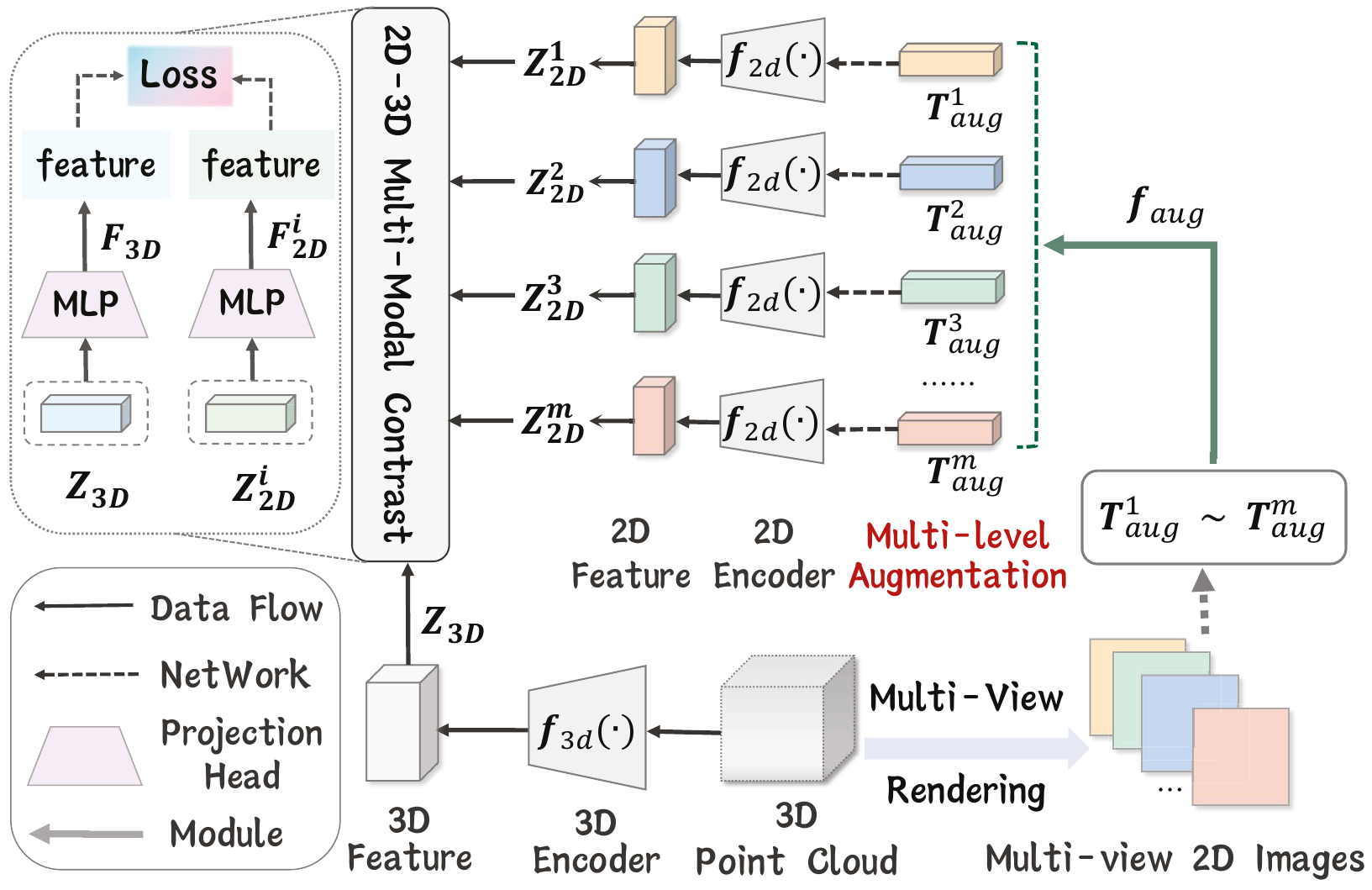}
  \caption{Schematic of Multi-level Augmentation. Multi-level augmentation  are generated in an incremental manner. 2D multi-views correspond to a certain level of augmentation indicated by different colors.}
  \label{fig:aug}
\end{figure}

An overview of our method is shown in Fig.\ref{fig:aug}. We propose an improvement to the contrastive framework for 3D point clouds and 2D multi-views through a multi-level augmentation module. Crucially, 
we employ an incremental strategy to generate multi-level augmentation, which in turn applies distinct transformations to the 2D views.

In 2D multi-view data, there are two types of information contained: the shared semantic information across all multi-views and the private information specific to each individual 2D view. If each view is treated equally with the same type or intensity of enhancement during training, the model will learn a non-optimal representation.

Meanwhile, in alignment with the \textit{InfoMax} \cite{bell1995information} principle, the goal of contrastive is to capture as much information as possible about stimuli. The \textit{InfoMin} \cite{noroozi2016unsupervised} principle suggests that further reduction of mutual information at the \textit{intermediate optimal} can be achieved by utilizing more robust augmentation.

Furthermore, we explore enhancing representational performance by mining hard samples through \textit{RandomCrop}. Given a 2D view image $I$, we first determine the cropping ratio $s$ and aspect ratio $r$ from a predefined range. 
This can be described as follows:

\begin{equation}
(x, y, h, w)=\mathbb{R}_{\textit {crop }}(s, r, I),
\end{equation}

where $\mathbb{R}_{\text {crop }}(\cdot, \cdot, \cdot)$ is a random sampling function that returns a quaternion $(x, y, h, w)$. where $(x, y)$ represents the coordinates of the cropping center, and $(h, w)$ represents the height and width of the cropping.

Assuming the number of 2D views is $\mathrm{\textit{m}}$, the complete augmentation pipeline is specified as 
$T=\textit{Combine}\left\{t_0, t_1, t_2, t_3, \cdots, t_m\right\}$ , where $t_0$ contains the basic augmentation and $t_1 \sim t_m$ 
represents a specific type of augmentation. Then, the incremental strategy can be represented as: 

\begin{equation}
\begin{gathered}
T_1=\textit{Combine}\left\{t_0, t_1\right\} \\
T_2=\textit{Combine}\left\{t_0, t_1, t_2\right\} \\
T_3=\textit{Combine}\left\{t_0, t_1, t_2, t_3\right\} \\
T_m=\textit{Combine}\left\{t_0, t_1, t_2, t_3, \cdots, t_m\right\}
\end{gathered}
\end{equation}

Using these modules, we transform the 2D multi-view samples $\left\{x_i\right\}_{1 \leq i \leq m}$ of the same 3D object into $m$ images with different augmentation intensities:

\begin{equation}
v_i=T_i\left(x_i\right), \quad i=1,2,3, \cdots, m
\end{equation}

The augmentation transformations from $T_1$ to $T_m$ gradually increase in intensity and number, reducing the shared mutual information between the transformed 2D view and the 3D object. Further, we use a projection head $g_i$ based on Multi-MLP to map the features into the loss space:
\begin{equation}
z_i=g_i\left(f_i\left(T_i\left(v_i\right)\right)\right), i=1,2,3, \cdots, m
\end{equation}

Here, $f_i$ represents the encoder and $z_i$ represents the features in the latent space. Note that the number of projection heads, rendered multi-view 2D images and augmentation modules are consistent. Also, the strength of the augmentation module and the variation trend of feature projection dimensions in Multi-MLP are consistent.

Therefore, the overall loss function formula is updated as follows, where $L_{\textit {contrast }}$ refers to the cross-modal loss between the 3D feature $z_{\mathrm{j}}$ and a specific 2D view feature $z_{\mathrm{i}}$: 
\begin{equation}
L_i=L_{\textit {contrast }}\left(z_i, z_j\right), \quad L_{\textit {overall }}=\sum_{i=1}^m L_i
\end{equation}

This way, the distribution of the augmentation invariance for $t_0$ and $t_1$ is the broadest, and the invariance of $t_m$ is limited to the corresponding features of the largest dimension.

\begin{table}[t]
  \centering
  \resizebox{0.98\linewidth}{!}{
  \Huge
    \begin{tabular}{cclcccc}
    \toprule
    \multicolumn{1}{c}{\multirow{2}[4]{*}{Type}} &       & \multicolumn{1}{c}{\multirow{2}[4]{*}{Method}} &       & \multicolumn{3}{c}{Accuracy (\%)} \\
\cmidrule{5-7}          &       &       &       & ModelNet40 &       & ModelNet10 \\
\cmidrule{1-1}\cmidrule{3-3}\cmidrule{5-5}\cmidrule{7-7}    \multirow{3}[2]{*}{\textit{Sup.}} &       & PointNet \cite{qi2017pointnet} &       & 89.2  &       & - \\
          &       & GIFT \cite{dovrat2019gift}  &       & 89.5  &       & 91.5  \\
          &       & MVCNN \cite{su2015multi}  &       & 89.7  &       & - \\
    \midrule
    \multirow{11}[4]{*}{\textit{Self.}} &       & Point-BERT \cite{yu2022point} &       & 87.4  &       & - \\
          &       & Point-MAE \cite{pang2022masked}  &       & 91.0  &       & - \\
\cmidrule{3-7}          
          &       & Jigsaw3D \cite{sauder2019self} &       & 90.6  &       & \underline{94.5}  \\
          &       & Vconv-DAE \cite{dovrat2021vconvdae}  &       & 75.5  &       & 80.5  \\
          &       & SwAV \cite{caron2020unsupervised}   &       & 90.3  &       & 93.5  \\
          &       & OcCo \cite{wang2020occ}   &       & 89.2  &       & 92.7  \\
          &       & STRL \cite{liu2021stochastic}  &       & 90.9  &       & - \\
          &       & CrossPoint \cite{afham2022crosspoint} &       & \underline{91.2}  &       & - \\
          &       & \cellcolor[rgb]{ .902,  .902,  .902}\textbf{MM-Point (ours)} & \cellcolor[rgb]{ .902,  .902,  .902} & \cellcolor[rgb]{ .902,  .902,  .902}\textbf{92.4 } & \cellcolor[rgb]{ .902,  .902,  .902} & \cellcolor[rgb]{.902,  .902,  .902}\textbf{95.4 } \\
          &       & \textit{Improvement} &       & \textbf{+1.2} &       & \textbf{+0.9} \\
    \bottomrule
    \end{tabular}%
    }
    \caption{Linear SVM classification results on ModelNet40 and ModelNet10. \textit{Self.} and \textit{Sup.} represent pre-training with self-supervised and supervised methods.}
  \label{tab:tab1}%
\end{table}%

\begin{figure*}[t]
  \centering
  \includegraphics[width=0.87\textwidth]{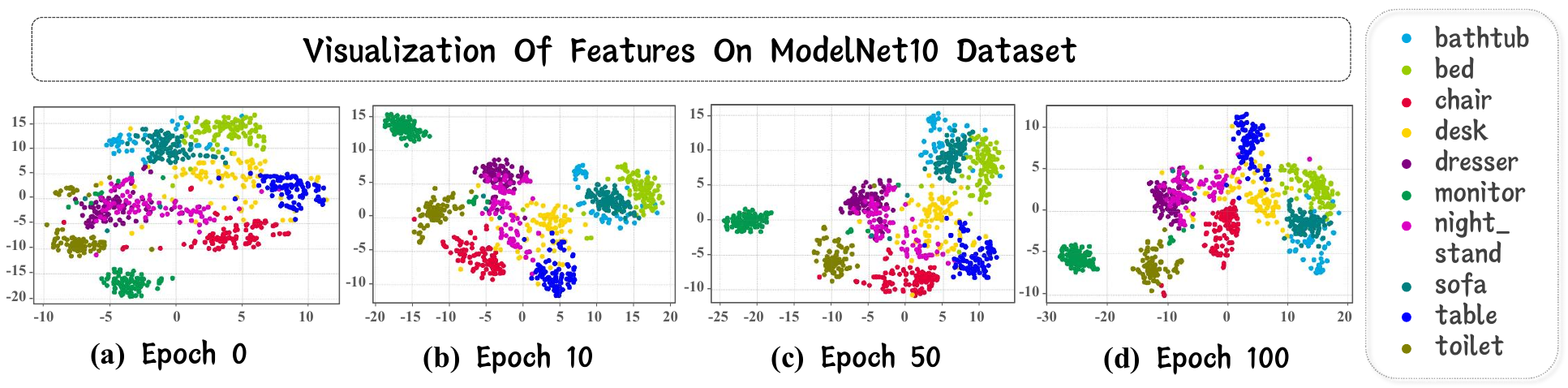}
  \caption{t-SNE visualization of point-level features extracted by MM-Point on ModelNet10 \cite{qi2016volumetric}, with each feature point colored according to its class label. Points with the same color represent semantic similarity. } 
  \label{fig:tsne}
\end{figure*}

\section{Experiments}
In this section, we first introduce the pre-training details of MM-Point. As our focus is on 3D representation learning, we only evaluate the pre-trained 3D point cloud encoder backbones. We sample different downstream tasks and assess the 3D feature representations learned by MM-Point. 

\subsection{Pre-training Setup}
\subsubsection{Datasets}
\textbf{ShapeNet} \cite{chang2015shapenet} is a large-scale 3D shape dataset containing $51162$ synthetic 3D point cloud objects.
For each object in the dataset, we render the 3D object into 2D multi-views, obtaining $24$ images per object.
\subsubsection{Implementation Details}
For the point cloud modality, we employ DGCNN \cite{wang2019dynamic} as the 3D backbone. For the image modality, we use ResNet-50 as the 2D backbone. For all encoders, we append a $2$-layer non-linear MLP projection head to generate the final representation. Note that we add different projection heads to obtain features.  
Pre-training employs AdamW as the optimizer.

\begin{table}[t]
  \centering
  \resizebox{0.85\linewidth}{!}{
  \Huge
    \begin{tabular}{lcc}
    \toprule
    \multicolumn{1}{c}{Method} &       & Accuracy (\%) \\
\cmidrule{1-1}\cmidrule{3-3}    
    GBNe \cite{touvron2020training} &       & 80.5  \\
    PRANet \cite{zhang2020simple} &       & 81.0  \\
    PointMLP \cite{ma2022rethinking} &       & 85.2  \\
    \midrule
    Point-BERT \cite{yu2022point} &       & 83.1  \\
    MaskPoint \cite{liu2022masked} &       & 84.3  \\
    Point-MAE \cite{pang2022masked} &       & 85.2  \\
    \midrule
    OcCo \cite{wang2020occ}   &       & 78.3  \\
    STRL \cite{liu2021stochastic}  &       & 77.9  \\
    CrossPoint \cite{afham2022crosspoint} &       & \underline{81.7}  \\
    \rowcolor[rgb]{ .902,  .902,  .902} \textbf{MM-Point (ours)} &       & \textbf{87.8 } \\
    \textit{Improvement} &       & \textbf{+6.1} \\
    \bottomrule
    \end{tabular}%
    }
    \caption{Evaluation of 3D point cloud linear SVM classification on ScanObjectNN.}
  \label{tab:tab2}%
\end{table}%

\subsection{3D Object Classification}
The point cloud classification experiments were conducted on three datasets.
Notably, we utilized a pre-trained encoder with frozen weights for evaluation using a linear SVM.

\textbf{ModelNet40} \cite{wu20153d} is a synthetic point cloud dataset obtained by sampling 3D CAD models, consisting of $12311$ 3D objects. 
\textbf{ModelNet10} \cite{qi2016volumetric} includes $4899$ CAD models with orientations from $10$ categories. \textbf{ScanObjectNN} \cite{uy2019revisiting} is a real-world 3D object dataset comprising $2880$ unique point cloud objects. 
This dataset offers a more realistic and challenging setting.

To evaluate the effectiveness of the point cloud representation learned by MM-Point, we first performed random sampling of $1024$ points for each object. 

The classification accuracy results for ModelNet40 and ModelNet10 are shown in Tab.\ref{tab:tab1}. As illustrated, MM-Point outperforms all existing self-supervised methods, achieving classification accuracies of $92.4\%$ and $95.4\%$, respectively.

To validate the effectiveness in the real world, we conducted experiments on ScanObjectNN, and the evaluation results are presented in Tab.\ref{tab:tab2}. In comparison with the state-of-the-art methods, the accuracy has been significantly improved by $6.1\%$, highlighting the advantages of MM-Point in challenging scenarios in real-world environments.

To visualize the learned 3D representations, we employ t-SNE to reduce the dimensionality of the latent representations and map them onto a 2D plane. Fig.\ref{fig:tsne} presents the visualization of 3D point cloud features learned by MM-Point.

\begin{table}[t]
  \centering
  \resizebox{0.99\linewidth}{!}{
  \Huge
    \begin{tabular}{lrccccc}
    \toprule
    \multicolumn{1}{c}{\multirow{2}[4]{*}{Method}} &       & \multicolumn{2}{c}{5-way} &       & \multicolumn{2}{c}{10-way} \\
\cmidrule{3-4}\cmidrule{6-7}          &       & 10-shot & 20-shot &       & 10-shot & 20-shot \\
    \midrule
          &       & \multicolumn{5}{c}{Results on ModelNet40 } \\
\cmidrule{1-1}\cmidrule{3-7}    3D-GAN \cite{wu2016learning} &       & 55.8\huge{±3.4} & 65.8\huge{±3.1} &       & 40.3\huge{±2.1} & 48.4\huge{±1.8}  \\
    PointCNN \cite{li2018pointcnn} &       & 65.4\huge{±2.8} & 68.6\huge{±2.2} &       & 46.6\huge{±1.5} & 50.0\huge{±2.3}  \\
    RSCNN \cite{li2018point} &       & 65.4\huge{±8.9} & 68.6\huge{±7.0} &       & 46.6\huge{±4.8} & 50.0\huge{±7.2} \\
    \midrule
    Jigsaw \cite{sauder2019self} &       & 34.3\huge{±1.3} & 42.2\huge{±3.5} &       & 26.0\huge{±2.4} & 29.9\huge{±2.6} \\
    OcCo \cite{wang2020occ}   &       & 90.6\huge{±2.8} & 92.5\huge{±1.9} &       & 82.9\huge{±1.3} & 86.5\huge{±2.2} \\
    CrossPoint \shortcite{afham2022crosspoint} &       & 92.5\huge{±3.0} & 94.9\huge{±2.1} &       & 83.6\huge{±5.3} & 87.9\huge{±4.2} \\
    \rowcolor[rgb]{ .902,  .902,  .902} \textbf{MM-Point (ours)} &       & \textbf{96.5\huge{±2.8}} & \textbf{97.2\huge{±1.4}} &       & \textbf{90.3\huge{±2.1}} & \textbf{94.1\huge{±1.9}} \\
    \midrule
          &       & \multicolumn{5}{c}{Results on ScanObjectNN } \\
\cmidrule{1-1}\cmidrule{3-7}    
    Jigsaw \cite{sauder2019self} &       & 65.2\huge{±3.8} & 72.2\huge{±2.7} &       & 45.6\huge{±3.1} & 48.2\huge{±2.8} \\
    OcCo \cite{wang2020occ}   &       & 72.4\huge{±1.4} & 77.2\huge{±1.4} &       & 57.0\huge{±1.3} & 61.6\huge{±1.2}  \\
    CrossPoint \shortcite{afham2022crosspoint} &       & 74.8\huge{±1.5} & 79.0\huge{±1.2} &       & 62.9\huge{±1.7} & 73.9\huge{±2.2} \\
    \rowcolor[rgb]{  .902,  .902,  .902} \textbf{MM-Point (ours)} &       & \textbf{88.0\huge{±6.5}} & \textbf{90.7\huge{±5.1}} &       & \textbf{76.7\huge{±1.9}} & \textbf{83.9\huge{±4.2}} \\
    \bottomrule
    \end{tabular}%
    }
    \caption{Few-shot classification: SVM classification accuracy comparison on ModelNet40 and ScanObjectNN. We report the average accuracy (\%) and standard deviation (\%).}
  \label{tab:tab3}%
\end{table}%

\subsection{3D Object Few-shot Classification }
The conventional setting for FSL is \textit{N-way K-shot}. We conduct experiments using the ModelNet40 and ScanObjectNN datasets. Specifically, we report the results of $10$ runs and calculate their mean and standard deviation.

We report the results in Tab.\ref{tab:tab3}. 
On both ModelNet40 and ScanObjectNN, MM-Point achieves \textbf{\textit{SOTA}} accuracy, outperforming other few-shot classification methods, and demonstrating substantial improvements across all settings.

\subsection{3D Object Part Segmentation}

We also extend MM-Point to the task of 3D object part segmentation, a challenging fine-grained 3D recognition task. 

The \textbf{ShapeNetPart} \cite{yi2016scalable}  dataset contains $16881$ point clouds, with 3D objects divided into $16$ categories and $50$ annotated parts. We sample $2048$ points from each input instance.

For a fair comparison, we follow previous works and add a simple part segmentation head on top of the DGCNN encoder. 
We evaluate the performance using Overall Accuracy (OA) and mean Intersection over Union (mIoU) metrics. Tab.\ref{tab:tab4} summarizes the evaluation results.

\begin{table}[t]
  \centering
  
  \resizebox{\linewidth}{!}{
  \Huge
    \begin{tabular}{crlrccc}
    \toprule
    \multirow{2}[4]{*}{Category} &       & \multicolumn{1}{c}{\multirow{2}[4]{*}{Method}} &       & \multicolumn{3}{c}{Metrics} \\
\cmidrule{5-7}          &       &       &       & OA (\%) &       & mIoU (\%) \\
\cmidrule{1-1}\cmidrule{3-3}\cmidrule{5-5}\cmidrule{7-7}    \multirow{3}[2]{*}{\textit{Sup.}} &       & PointNet \cite{qi2017pointnet} &       & -     &       & 83.7  \\
          &       & PointNet++ \cite{qi2017pointnet++} &       & -     &       & 85.1  \\
          &       & DGCNN \cite{wang2019dynamic} &       & -     &       & 85.1  \\
    \midrule
    \multirow{3}[2]{*}{\textit{Self-sup.}} 
          &       & OcCo \cite{wang2020occ}   &       & 94.4  &       & 85.0  \\
          &       & CrossPoint \cite{afham2022crosspoint} &       & \underline{94.4}  &       & \underline{85.3}  \\
          &       & \cellcolor[rgb]{ .902,  .902,  .902}\textbf{MM-Point (ours)} & \cellcolor[rgb]{ .902,  .902,  .902} & \cellcolor[rgb]{ .902,  .902,  .902}\textbf{94.5} & \cellcolor[rgb]{ .902,  .902,  .902} & \cellcolor[rgb]{ .902,  .902,  .902}\textbf{85.7} \\
    \bottomrule
    \end{tabular}%
    }
    \caption{Overall accuracy and mean IoU results for 3D part segmentation. The metrics are OA(\%) and mIoU(\%).}
  \label{tab:tab4}%
\end{table}%

\subsection{3D Object Semantic Segmentation}
The Stanford Large-Scale 3D Indoor Spaces Dataset (\textbf{S3DIS}) \cite{armeni20163d} consists of 3D scan data from $271$ rooms across six different indoor spaces.
For evaluation, we train our model from scratch on Areas $1\sim4$ and Area $6$, using Area $5$ for validation.

Tab.\ref{tab:tab5} demonstrates the performance of MM-Point. Compared to the randomly-initialized baseline without pre-training, our proposed MM-Point pre-training method yields a significant improvement ($+4.2\%$ mIoU).

\begin{table}[t]
  \centering
  
  \resizebox{0.84\linewidth}{!}{
  \Huge
    \begin{tabular}{lcccc}
    \toprule
    \multicolumn{1}{c}{\multirow{2}[4]{*}{Method}} &       & \multicolumn{3}{c}{Metrics} \\
\cmidrule{3-5}          &       & OA(\%) &       & mIoU(\%) \\
\cmidrule{1-1}\cmidrule{3-3}\cmidrule{5-5}    
    Jigsaw3D \cite{sauder2019self} &       & 84.1  &       & 55.6 \\
    OcCo \cite{wang2020occ}   &       & 84.6  &       & 58 \\
    CrossPoint \cite{afham2022crosspoint} &       & \underline{86.7}  &       & \underline{58.4} \\
    \rowcolor[rgb]{ .902,  .902,  .902} \textbf{MM-Point (ours)} &       & \textbf{88.7} &       & \textbf{59.1} \\
    \bottomrule
    \end{tabular}%
    }
    \caption{Semantic segmentation results on S3DIS \cite{armeni20163d}. We report the \textit{mIoU} and \textit{OA} for all 13 classes.}
  \label{tab:tab5}%
\end{table}%

\subsection{Ablation Study}
In order to investigate the contributions of each main component in MM-Point, we conduct an extensive ablation study. 
\subsubsection{Multi-view Contrastive: Number of Views}
We evaluate the performance of MM-Point by performing multi-modal contrastive with different numbers of 2D views. 
As shown in Tab.\ref{tab:tab6}, we observe that the performance of MM-Point is the lowest when using only one view image. As more 2D image views are added, the classification performance steadily improves. The performance is best when the number of multi-view 2D images $M$ is $4$.

\begin{table}[t]
  \centering
  \resizebox{\linewidth}{!}{
  \Huge
    \begin{tabular}{c|c|ccccccccccc}
    \toprule
    \multicolumn{2}{c|}{Number of 2D images} &        & 1     &       & 3     &       & \textbf{4} &       & 5     &       & 6 \\
    
    \midrule
    \multirow{2}[4]{*}{Accuracy (\%)} & ModelNet40 &        & 91.3  &       & 92.2  &       & \textbf{92.4} &       & 92.4  &       & 92.1\\
    
\cmidrule{2-3}\cmidrule{4-4}\cmidrule{6-6}\cmidrule{8-8}\cmidrule{10-10}\cmidrule{12-12}          & ScanObjectNN &       & 83.3  &       & 86.7  &       & \textbf{87.8} &       & 87.6  &       & 86.9\\
    
    \bottomrule
    \end{tabular}%
    }
    \caption{Ablation test using different numbers of 2D views.}
  \label{tab:tab6}%
\end{table}%

\subsubsection{Multi-MLP Strategy}
Employing different MLPs for \textit{intra-modal} and \textit{inter-modal} feature embedding, as well as for different dimensional embedding of multi-views, is a distinctive design in MM-Point. We evaluate our method by: (1) using a unified MLP, 
(2) employing different dimensions for \textit{intra-modal} and \textit{inter-modal} features, ensuring that \textit{inter-modal} are larger than \textit{intra-modal} dimensions, and (3) using unified output dimensions or multiple different dimensions for 2D multi-views. The results are reported in Tab.\ref{tab:tab7} . These results indicate that our unified framework benefits from using multiple different spaces for multi-modal modeling.

\begin{table}[t]
  \centering
  \resizebox{0.98\linewidth}{!}{
  \Huge
    \begin{tabular}{ccccccc}
    \toprule
    \multicolumn{3}{c}{Multi-MLP} &       & \multicolumn{3}{c}{Accuracy (\%)} \\
\cmidrule{1-3}\cmidrule{5-7}    Intra-modal &       & Inter-modal &       & ModelNet40 &       & ScanObjectNN \\
\cmidrule{1-1}\cmidrule{3-3}\cmidrule{5-5}\cmidrule{7-7}    \ding{55}     &       & \ding{55}     &       & 91.4  &       & 83.4 \\
    \ding{55}     &       & \ding{52}     &       & 91.9  &       & 86.7 \\
    \ding{52}     &       & \ding{55}     &       & 91.6  &       & 84.9 \\
    \ding{52}     &       & \ding{52}     &       & \textbf{92.4} &       & \textbf{87.8} \\
    \bottomrule
    \end{tabular}%
    }
    \caption{A comparison using different Multi-MLP strategies.}
  \label{tab:tab7}%
\end{table}%

\subsubsection{Multi-level Augmentation Strategy}
To validate the effectiveness of the multi-level augmentation strategy, we experiment with: (1) all 2D views based solely on unified augmentations, (2) all 2D views based on different augmentations, but without increasing the difficulty in a hierarchical manner, and (3) all 2D views based on different multi-level augmentations. The results are reported in Tab.\ref{tab:tab8}. 
The absence of any specific augmentation (indicated by \ding{55} ) negatively impacts the performance. This suggests that applying multi-level augmentations to 2D multi-views allows the model to benefit from contrast in the augmentation space.

\begin{table}[h]
  \centering
  
  \resizebox{0.97\linewidth}{!}{
  \Huge
    \begin{tabular}{ccccccc}
    \toprule
    \multicolumn{3}{c}{Multi-level Augmentation} &       & \multicolumn{3}{c}{Acc. (\%)} \\
\cmidrule{1-3}\cmidrule{5-7}    Multi Aug. &       & Multi-level &       & ModelNet40 &       & ScanObjectNN \\
\cmidrule{1-1}\cmidrule{3-3}\cmidrule{5-5}\cmidrule{7-7}    \ding{55}     &       & \ding{55}     &       & 91.7  &       & 86.1 \\
    \ding{52}     &       & \ding{55}     &       & 92.1  &       & 86.9 \\
    \ding{52}     &       & \ding{52}    &       & \textbf{92.4} &       & \textbf{87.8} \\
    \bottomrule
   \end{tabular}%
    }
    \caption{The impact of Multi-level augmentation strategy.}
  \label{tab:tab8}%
\end{table}%

\section{Conclusion}
In this paper, we explore a novel pre-training method for 3D representation learning.
We introduce MM-Point, a framework that encourages 3D point clouds to learn excellent features from 2D multi-views. Concurrently, MM-Point employs Multi-MLP and Multi-level augmentation strategies to effectively learn more robust 3D modality knowledge from 2D multi-views. MM-Point consistently exhibits state-of-the-art performance on various downstream tasks.
Codes are available at \url{https://github.com/HaydenYu/MM-Point}.

\section{Acknowledgments}
This work was supported by National Natural Science Foundation of China $61906036$ and the Fundamental Research Funds for the Central Universities $(2242023k30051)$.
This research work was also supported by the Big Data Computing Center of Southeast University

\bibliography{aaai24}

\end{document}